\documentclass[letterpaper, 10 pt, conference]{ieeeconf}  

\IEEEoverridecommandlockouts                              

\usepackage{graphics} 
\usepackage{epsfig} 
\usepackage{mathptmx} 
\usepackage{times} 
\usepackage{amsmath} 
\usepackage{hyperref}
\hypersetup{
  colorlinks = true, 
  urlcolor = black, 
  linkcolor = black, 
  citecolor = black 
}
\usepackage{multirow}
\usepackage{graphicx}

\title{\LARGE \bf
Developing a Robotic Surgery Training System for Wide Accessibility and Research*
}

\author{Walid Shaker and Mustafa Suphi Erden
\thanks{*This study was partially funded by the Engineering and Physical Sciences Research Council of the UK, under the EPSRC Grant EP/Y017307/1.}
\thanks{Walid Shaker and Mustafa Suphi Erden are with the School of Engineering and Physical Sciences, Heriot-Watt University, Edinburgh EH14 4AS, UK
{\tt\small (wkhs2000@hw.ac.uk; m.s.erden@hw.ac.uk)}}
\thanks{Walid Shaker is the corresponding author of this study {\tt\small(+447747370781; wkhs2000@hw.ac.uk)}}
}

\begin{document}
\maketitle
\thispagestyle{empty}
\pagestyle{empty}

\begin{abstract}

Robotic surgery represents a major breakthrough in medical interventions, which has revolutionized surgical procedures. However, the high cost and limited accessibility of robotic surgery systems pose significant challenges for training purposes. This study addresses these issues by developing a cost-effective robotic laparoscopy training system that closely replicates advanced robotic surgery setups to ensure broad access for both on-site and remote users. Key innovations include the design of a low-cost robotic end-effector that effectively mimics high-end laparoscopic instruments. Additionally, a digital twin platform was established, facilitating detailed simulation, testing, and real-time monitoring, which enhances both system development and deployment. Furthermore, teleoperation control was optimized, leading to improved trajectory tracking while maintaining remote center of motion (RCM) constraint, with a RMSE of 5 $\mu$m and reduced system latency to 0.01 seconds. As a result, the system provides smooth, continuous motion and incorporates essential safety features, making it a highly effective tool for laparoscopic training.

\end{abstract}

\section{INTRODUCTION}

Robot-assisted surgery has been increasingly expanding in adoption due to its numerous advantages over open surgery and traditional laparoscopic minimally-invasive surgery (MIS) \cite{zidane2023robotics}. This pioneering surgical procedure offers a plethora of benefits for both patients and surgeons. Similar to MIS, it utilizes small incisions, made by trocars to insert a laparoscope and surgical tools, leading to reduced patient pain, minimal scarring, and faster recovery \cite{lee2018master}-\cite{girerd2020design}, thereby alleviating stress on the immune system and shortening hospital stays \cite{meinzer2020advances}. Surgeons, meanwhile, benefit from improved visual fields and superior robot dexterity \cite{barua2024innovations}. However, surgeons must undergo rigorous training and continuous education to develop specialized skills and adapt to the demands of robotic surgery.

Despite the critical need for training, accessible and affordable robotic surgery training platforms are lacking. Surgeons typically perform training on the same systems used in real surgeries, requiring institutions to invest heavily in separate training setups. This proves costly for institutions, restricts surgeon access to training, and severely limits opportunities for researchers to experiment \cite{sridhar2017training}. Therefore, these high costs and inadequate training for surgeons to operate these sophisticated systems are the most prominent challenges, which are limiting the adoption of these systems \cite{meinzer2020advances}. 

To tackle these issues, our laboratory developed a low-cost experimental robotic surgery setup: RoboScope, designed to imitate existing systems while promoting broader accessibility for training and research purposes. 

\section{BACKGROUND AND LITERATURE REVIEW}

While robot-assisted surgery continues to expand into various surgical procedures, there are growing concerns regarding its higher costs \cite{sheetz2020trends}. Systems like da Vinci\footnote{\href{https://www.intuitive.com/en-us/patients/da-vinci-robotic-surgery/}{https://www.intuitive.com/en-us/patients/da-vinci-robotic-surgery/}} by Intuitive Surgical (Sunnyvale, CA, USA) and Versius\footnote{\href{https://cmrsurgical.com/versius/}{https://cmrsurgical.com/versius/}} by CMR Surgical (Cambridge, UK) are notably expensive for hospitals to procure. For example, the widely adopted da Vinci robotic surgery system can cost up to $\$2.5$ million, and additional annual maintenance fees of around $\$190,000$ \cite{eckhoff2023costs}. In addition, these systems are primarily utilized by experienced surgeons, imposing time constraints on trainees and limiting their opportunities for comprehensive training. 

Notably, our system design solution comes at a significantly lower cost, priced at approximately £60,000, making it a more accessible option for educational and research institutions. 

\subsection{System Preliminary Version}

\begin{figure}[t!]
\centering
\includegraphics[scale=0.41]{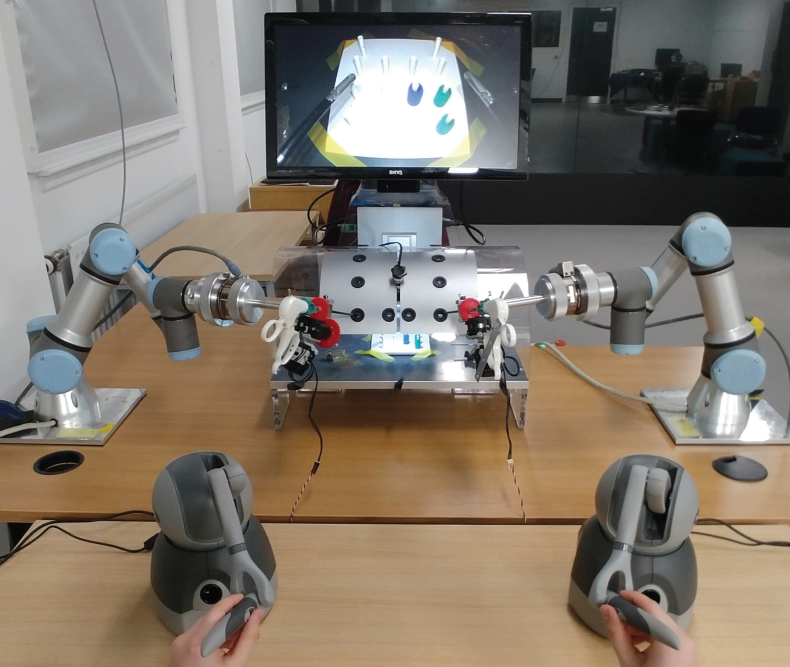}
\caption{Overview of the preliminary setup for the robotic surgery trainer, illustrating the key components and configuration \cite{trute2022development}.}
\label{fig:Old_system}
\end{figure}

The preliminary version of the robotic surgery trainer, as we described in \cite{trute2022development}, prioritizes widespread deployability to ensure broad accessibility among surgical trainees. This objective is achieved by constructing the system with readily available, off-the-shelf components, making it affordable for standard engineering laboratories. The system includes key components such as a laparoscopy training box, two UR3 robotic arms with force/torque sensors, and standard laparoscopic instruments controlled by servo motors via an Arduino processor. The system supports teleoperation through two Touch haptic devices, while visual feedback is provided either via a fisheye lens camera or through two Sony cameras with an Oculus VR headset. The setup also features a laptop for system management and is supported by two software applications designed for on-site and off-site control. The preliminary setup is shown in Fig.~\ref{fig:Old_system}.

\hfill
\subsubsection{Robotic Arm} Two UR3 CB3 robotic arms are employed to execute motion commands provided by the user to maneuver the laparoscopic tools within the training box. The selection of UR3 robots was made based on their compact design, which suits this application within a small and confined workspace. Yet, these robots can access all areas required inside the box for typical suturing activities as the workspace of the laparoscopy training box is represented by a rectangular prism measuring 30x20x15 cm. These collaborative robots feature back-drivable joints, allowing users to manually guide the robot and establish the initial configuration with the laparoscopic instrument. While UR3 robots offer safety advantages in case of hard collisions compared to industrial counterparts \cite{arents2021human}, their repeatability precision may be slightly compromised. However, this difference in accuracy is not critical for this application.

\hfill
\subsubsection{Preliminary End-effector Design} In the preliminary version of the setup \cite{trute2022development}, we equipped each robot wrist with a force/torque sensor to detect contact with the laparoscopic tool tip and provide estimates of both the magnitude and direction of the contact force. The sensor is followed by an attachment designed to hold the laparoscopic tool. 

\begin{figure}[t!]
\centering
\includegraphics[scale=0.25]{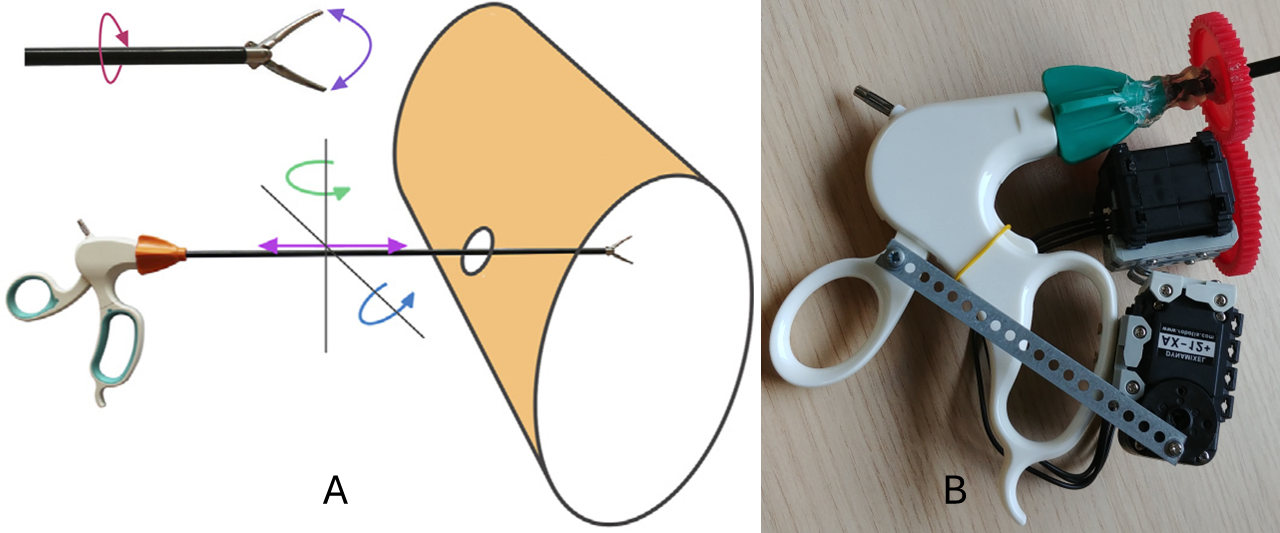}
\caption{(A) The actuated DOF of the laparoscopic tool. (B) The modifications made to the forceps for controlling the forceps head axial rotation and grasping motion using two Dynamixel servos \cite{trute2022development}.}
\label{fig:Old_DOF_forceps}
\end{figure}

The preliminary design utilized a laparoscopic instrument, specifically forceps, which was fully controlled in five degrees of freedom (DOF) and constrained by a hole in the training box, as shown in Fig.\ref{fig:Old_DOF_forceps}a. Three of these DOFs were managed by the robot arm to adjust the position of the tool tip, which includes two rotational movements around orthogonal axes and one translational movement for inserting or retracting the tool through the hole. The fourth DOF, enabling continuous axial rotation of the forceps, was controlled by a servo gear mechanism attached to the instrument. The fifth DOF, which controls the opening and closing of the tool gripper, was managed by a separate servo within predefined limits for the fully open and fully closed positions. This mechanism employed a thin metal strip that connects the servo horn to the adjustable thumb handle of the instrument. The preliminary setup of the laparoscopic instrument, including the two servo mounts, is illustrated in Fig.~\ref{fig:Old_DOF_forceps}b.

\subsection{System Limitations}

The preliminary setup has faced several limitations, such as delays in robot response, occasional non-smooth motion, slow robot movements, and inadequate safety features. Additionally, the end-effector design had flaws, including suboptimal adjustments to the laparoscopic instrument and its insecure mounting. These factors posed significant challenges to the system's overall reliability and robustness.

The current study aims to further develop the system by mitigating the above-mentioned limitations. The teleoperation control has been improved to eliminate delays in robot response and ensure smooth, continuous, and notably faster robot motion. Moreover, an innovative design of the end-effector has been implemented to address the preliminary design drawbacks. Additionally, a dedicated digital twin has been established for rigorous experimentation and validation, providing a controlled environment to fine-tune advancements before real-world deployment. This digital twin facilitates remote operations by providing real-time monitoring and visualization of the robot's movements.


\section{SYSTEM DEVELOPMENT}

\begin{figure}[t!]
\centering
\includegraphics[scale=0.088]{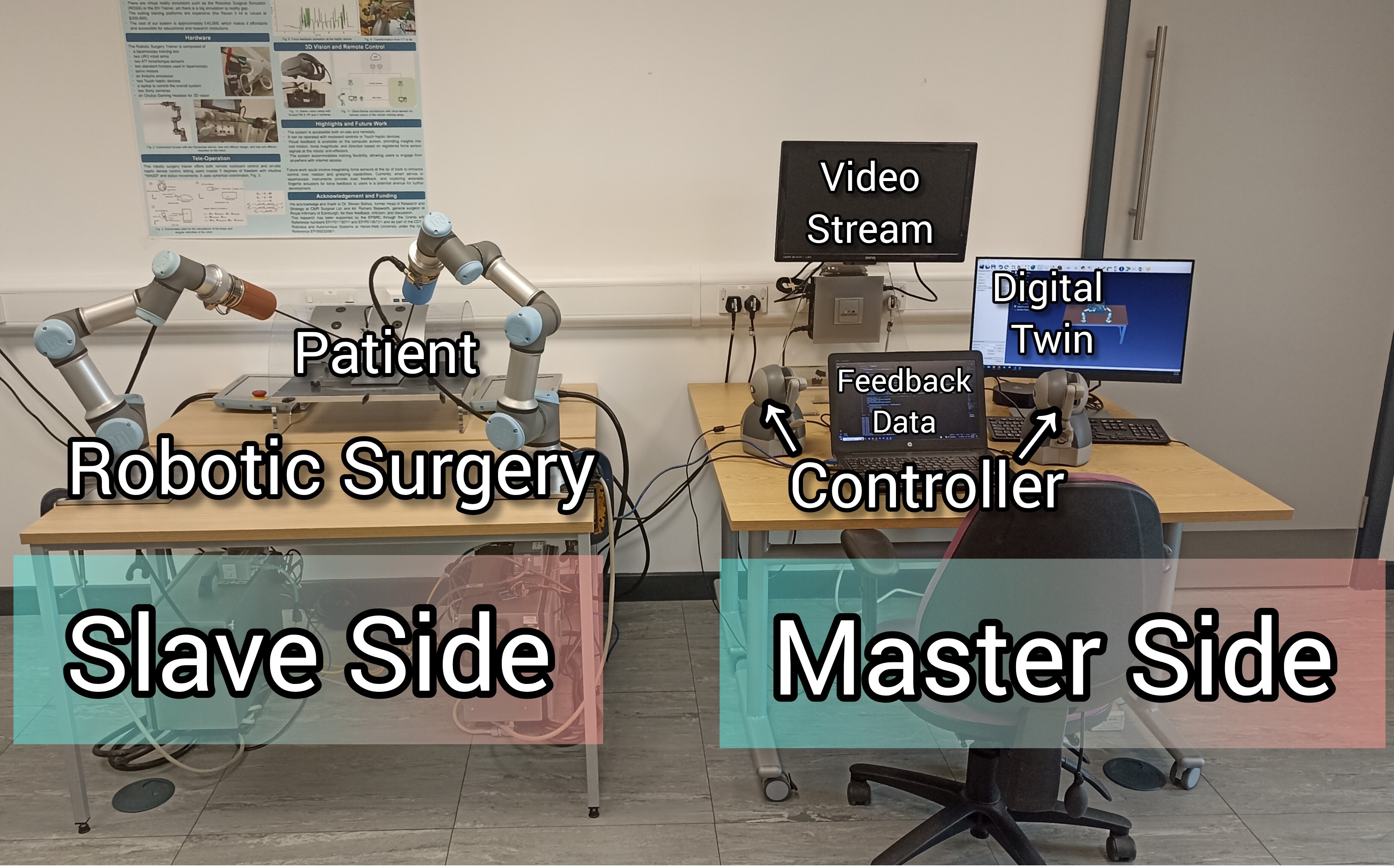}
\caption{System architecture of the developed master-slave robotic surgery training setup (RoboScope).}
\label{fig:Master-Slave2}
\end{figure}

The system has been reconfigured, and a master-slave robotic surgery training system has been developed to address the challenges of the initial version. This new setup supports the execution of various standard training exercises by enabling the teleoperation of two laparoscopic instruments within a training box. The box includes holes that simulate incisions or transabdominal working ports on patients. The overall system architecture is depicted in Fig.~\ref{fig:Master-Slave2}.

The system operates in two distinct modes: on-site and off-site. In on-site mode, users can control the system using either a keyboard or Touch haptic devices, which provide haptic feedback by capturing interaction forces at the instrument tip. Off-site mode enables remote teleoperation through personal computers connected to the internet, using keyboard input. On-site users can choose between 2D or 3D visual feedback. The 2D visual feedback is delivered via a fisheye lens camera, with real-time tool movements displayed on a monitor at the master station. For enhanced depth perception, 3D visual feedback is available through two wide-angle cameras and an Oculus Rift S VR headset, offering stereo vision. Remote users have access to a live video stream that shows the inside of the training box, as well as a digital twin software that replicates the workspace environment. An on-site supervisor is present to ensure proper setup and maintain system integrity during remote operations.

This section outlines the key improvements made to the system, including the design of a new end-effector and the development of a digital twin platform, while details on the enhancement of teleoperation control are provided in Section.~\ref{sec:teleop}.

\subsection{New End-effector Design}

\begin{figure}[t!]
\centering
\includegraphics[scale=0.29]{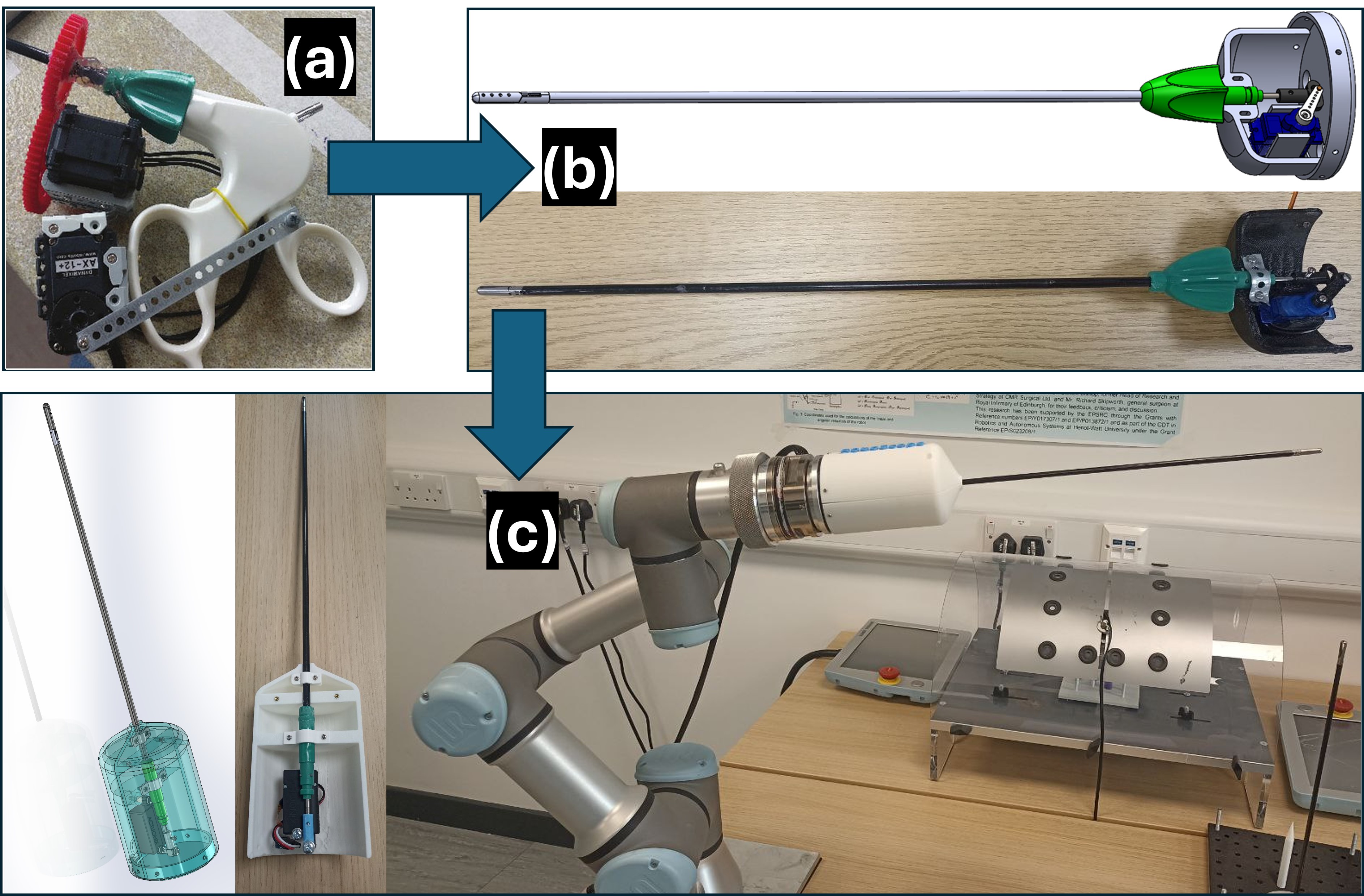}
\caption{Development stages of the end-effector: (A) Initial design; (B) Second iteration, which suffers from bending at the free end of the forceps head; (C) Final optimized version.}
\label{fig:EE_development}
\end{figure}

A new robot end-effector design has been introduced to address the shortcomings of the previous iteration. Utilizing CAD software and adhering to Design for Manufacturing (DFM) and Design for Assembly (DFA) principles, the new design closely emulates fully-fledged robotic surgery systems. The development of the end-effector, specifically the laparoscopic instrument, has undergone three iterations, as illustrated in Fig.~\ref{fig:EE_development}. 

In contrast to the preliminary design, where the tool lacked proper fixation and was merely inserted into a hole, the second iteration significantly improves stability and securely mounts the laparoscopic instrument onto the robot. This improvement simplifies the adjustments required for the forceps, shown in Fig.\ref{fig:Old_DOF_forceps}b, by eliminating the servo motor that controls the fourth DOF, which is responsible for the axial rotation of the forceps head. Instead, this DOF is now controlled by the robot's $6^{th}$ joint in a hold-and-release mode, allowing for continuous control of both clockwise and counterclockwise rotations. Additionally, the fifth DOF, which controls the opening and closing of the forceps, is managed by a servo gripping mechanism built inside the encapsulating box of the instrument, as depicted in Fig.~\ref{fig:EE_development}b.

Despite these improvements, the second version exhibited cantilever-like behavior, resulting in deflection at the free end of the forceps' axial head. To address this issue, the final optimized version, shown in Fig.~\ref{fig:EE_development}c, solves this by adding more support at the fixed end and shortening the axial head to effectively reduce the deflection. 

To integrate the new design with the physical robot, it was essential to calculate the payload on the robot's wrist and calibrate the new Tool Center Point (TCP) at the tip of the forceps head. This calibration used a 4-point method, which aligns the tool tip at four different orientations around a single reference point, providing precise positioning and orientation of the tool tip relative to the robot's base. 

\subsection{Digital Twin Simulation and Testing}

\begin{figure}[t!]
\centering
\includegraphics[scale=0.47]{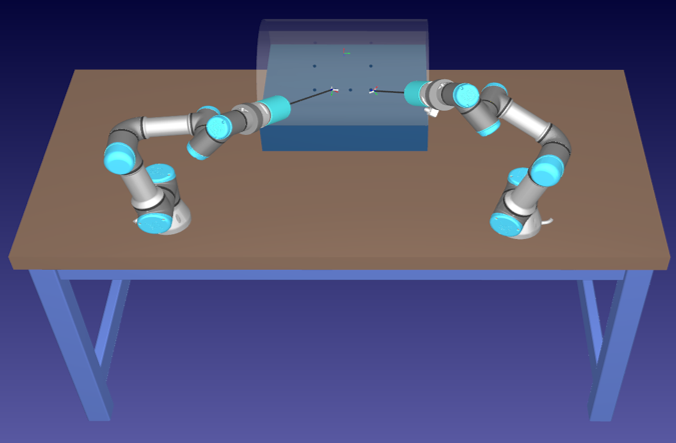}
\caption{Digital twin simulation environment developed using RoboDK.}
\label{fig:Digital_twin_env2}
\end{figure}

A digital twin was developed using RoboDK\footnote{\href{https://robodk.com/}{https://robodk.com/}}, an industrial robot programming and simulation software, and RoboDK API for Python. This platform plays a vital role in testing, validating, and refining system enhancements before real-world deployment. The digital twin introduces safety features by checking for singularities, joint limits, and collision avoidance, reducing risks and optimizing performance. The simulation environment includes a model of a table housing two UR3 robots and the laparoscopic training box, as shown in Fig.~\ref{fig:Digital_twin_env2}. The laparoscopic training box features transparent interiors for better visualization and multiple holes through which the instrument can pass to simulate incisions for laparoscopic procedures. This platform is especially valuable for remote users, as it combines simulation with real-time data to replicate robot movements visually. By continuously retrieving real robot positions, the virtual robots in RoboDK accurately mimic these movements, providing a real-time representation of the physical robots' actions.

\section{TELEOPERATION PROCESS}
\label{sec:teleop}

\begin{figure}[b!]
\centering
\includegraphics[scale=0.33]{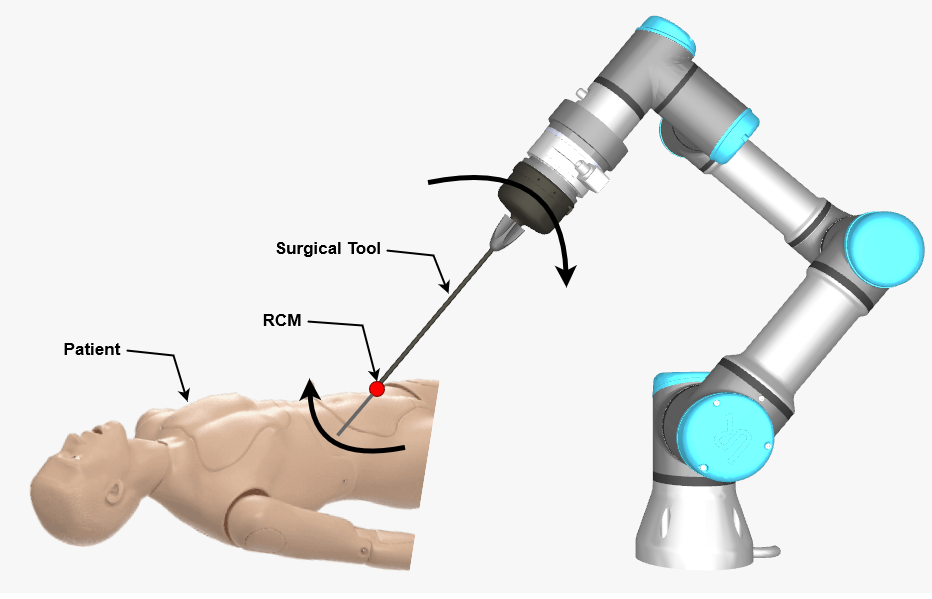}
\caption{Surgical tool pivoting around RCM insertion point.}
\label{fig:RCM}
\end{figure}

Since MIS is performed through small incisions in the patient's body, the surgical instrument must pivot around these points to avoid slipping away and damaging the abdominal walls \cite{colan2023concurrent}. This requirement is known as the Remote Center of Motion (RCM). In this system, a 5 mm diameter forceps head passes through an 8 mm diameter hole in the laparoscopy training box. Thus, it is crucial to restrict the movements of this tool to ensure that it consistently passes through the hole pivotal point without collisions. This pivot point, often referred to as the fulcrum point, poses a significant challenge in robotic surgery due to the need for accurate compensation. The fulcrum effect compensation requires that to move the tool tip in one direction, surgeons must move their hand in the opposite direction as the tool pivots around the incision, as illustrated in Fig.~\ref{fig:RCM}. However, the system successfully compensates for the fulcrum effect, facilitating smooth and accurate robotic surgery training.

\subsection{Remote Center of Motion}

\begin{figure}[b!]
\centering
\includegraphics[scale=0.15]{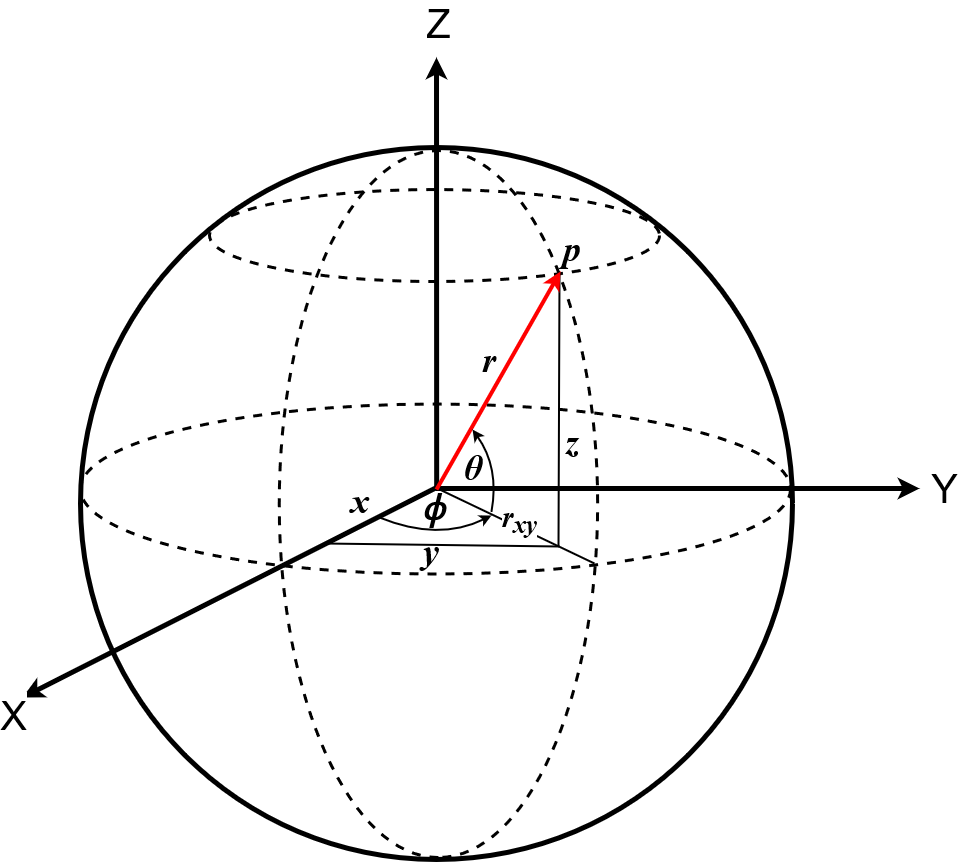}
\caption{Spherical coordinate system used to implement the RCM constraint.}
\label{fig:Sphere}
\end{figure}

The implementation of the RCM constraint reveals a similarity to the spherical coordinate system, where any point ($p$) is defined by longitude ($\phi$), latitude ($\theta$), and radius ($r$), as illustrated in Fig.~\ref{fig:Sphere}. Under this analogy, the RCM constraint can be conceptualized using spherical coordinates: the center of the sphere aligns with the center of the training box hole, and any point on the sphere ($p$) represents the position of the robot flange. The radius of the sphere corresponds to the length of the surgical tool outside of the training box.

The adjustments in spherical coordinates are described by\begin{align}
\begin{split}
\theta_{i+1} &= \theta_i + \Delta \theta \\
\phi_{i+1} &= \phi_i + \Delta \phi \\
r_{i+1} &= r_i + \Delta r \\
\end{split}
\label{eq:sph}
\end{align}

\noindent Subsequently, the position of the robot flange $(x^p, y^p, z^p)$ is computed relative to the recorded hole position $(x^h, y^h, z^h)$ as follows:
\begin{align}
\begin{split}
x_{i+1}^{p} &= x^h + r_{i+1} \cos(\theta_{i+1}) \cos(\phi_{i+1}) \\
y_{i+1}^{p} &= y^h + r_{i+1} \cos(\theta_{i+1}) \sin(\phi_{i+1}) \\
z_{i+1}^{p} &= z^h + r_{i+1} \sin(\theta_{i+1})
\end{split}
\label{eq:XYZ}
\end{align}

The orientation of the robot flange is characterized by the ZYX Euler angles representation, where changes in longitude and latitude correspond to adjustments in the roll and yaw angles of the robot flange, respectively. The proposed method simplifies this process by assuming a zero-pitch robot orientation. Therefore, at the program's start, the $6^{th}$ joint of the robot is rotated to ensure that the pitch angle is set to zero. During teleoperation, any rotation of the $6^{th}$ joint by the user to manipulate the forceps head is compensated for in the RCM calculation. Moreover, the program records the initial orientation ($roll_0, pitch_0, yaw_0$) alongside the starting longitude ($\phi_0$) and latitude ($\theta_0$) values. Thus, the orientation of the robot flange is computed as follows:

\begin{align}
\begin{split}
roll_i &= roll_0 +  (\theta_{i+1} - \theta_0) \\
pitch_i &= pitch_0 \\
yaw_i &= yaw_0 +  (\phi_{i+1} - \phi_0) \\
\end{split}
\label{eq:RPY}
\end{align}

To ensure smooth and continuous motion of the tool, velocity control is preferred over position control. To achieve this, a Jacobian matrix is utilized to relate the robot flange velocities to the robot joints velocities. The linear velocities of the robot flange $(\Delta x, \Delta y, \Delta z)$ are determined by constructing a vector from $p_i$ to $p_{i+1}$, while the angular velocities $(\omega x, \omega y, \omega z)$ are computed from the quaternion representation of the difference matrix between the two rotation matrices from $p_i$ to $p_{i+1}$. By constructing a robot flange velocity vector $v = [\Delta x, \Delta y, \Delta z, \omega x, \omega y, \omega z]^T$, the robot joint velocities $\dot{q}$ are obtained by 

\begin{equation}
    \dot{q} = J^{-1}v
\end{equation}

\noindent where $J$ is a $6\times6$ Jacobian matrix calculated numerically.

\subsection{Teleoperation Controllers}

To enable teleoperation, the system supports control via both keyboard and Touch haptic devices, which allows users to manipulate five degrees of freedom: three-dimensional linear movement of the tool tip, rotation of the forceps head, and grasping actions. The keyboard serves as a practical option for remote users, offering control from any computer with internet access. In contrast, the Touch devices offer a more user-friendly and ergonomic interface, specifically designed for on-site practice. The control scheme for the keyboard interface is detailed in Table~\ref{table:keyboard}, which outlines the commands used to operate both the left and right surgical instruments within the laparoscopic box.
    
\renewcommand{\arraystretch}{1.2}
\begin{table}[b!]
\centering
\caption{Keyboard Scheme for Surgical Tools Teleoperation.}
\begin{tabular}{ccc}
\hline 
\textbf{Tool Tip Movement} & \textbf{Keys (left)}  & \textbf{Keys (right)}\\
\hline 
In / Out & W / S & I / K \\
Left / Right & A / D & J / L \\
Up / Down & Q / E & U / O \\
Rotate CW / CCW & C / X & M / N \\
Grasp / Release & R / F & Y / H \\
Speed Increase / Decrease & LCtrl / LAlt & RCtrl / RAlt \\
\hline
\end{tabular}
\label{table:keyboard}
\end{table}

Two 3D Systems' Touch haptic devices are integrated to implement bilateral teleoperation. These devices offer 6 DOF for position sensing, including Cartesian coordinates $(X, Y, Z)$ and rotational angles $(Roll, Pitch, Yaw)$. However, they provide force feedback in three dimensions $(X, Y, Z)$, which is sufficient to feedback both the magnitude and direction of interaction forces upon detecting contact. 

For these devices, the motion of the stylus was mapped to control the tool tip's movement within the world frame. A dead zone was introduced for both linear and rotational motions to prevent the robot from responding to minor or unintended hand movements. Commands are only transmitted when the stylus moves beyond a defined threshold, ensuring that any small displacements are ignored. This minimizes unsteady movements of the tool tip within the training box due to hand tremors. 

Within the dead zone, the stylus buttons allow the user to control the laparoscopic tool's grasp and rotation functions. Pressing both buttons simultaneously temporarily halts command transmission, putting the stylus in a free state. This functionality allows the user to rest their hand or rotate the tool without affecting the robot’s actions, thereby enhancing both control and stability during teleoperation. In this work, the Touch haptic devices serve as the primary interface for teleoperation control.

\section{RESULTS}

\begin{figure}[t!]
\centering
\includegraphics[scale=0.7]{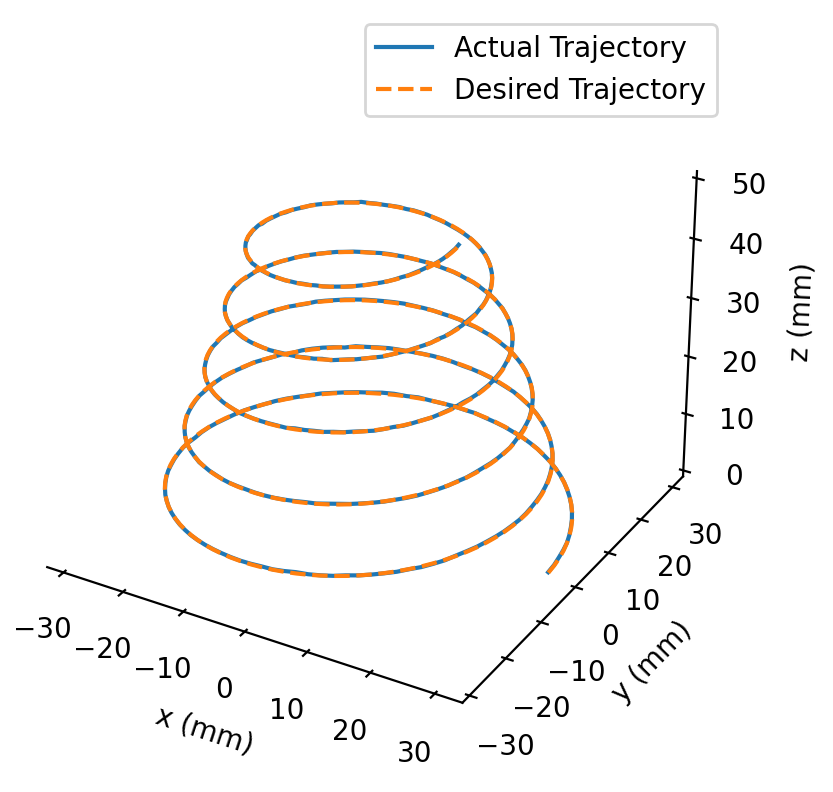}
\caption{Truncated cone tool tip trajectory with RCM constraint.}
\label{fig:cone}
\end{figure}

\begin{figure}[t!]
\centering
\includegraphics[scale=0.7]{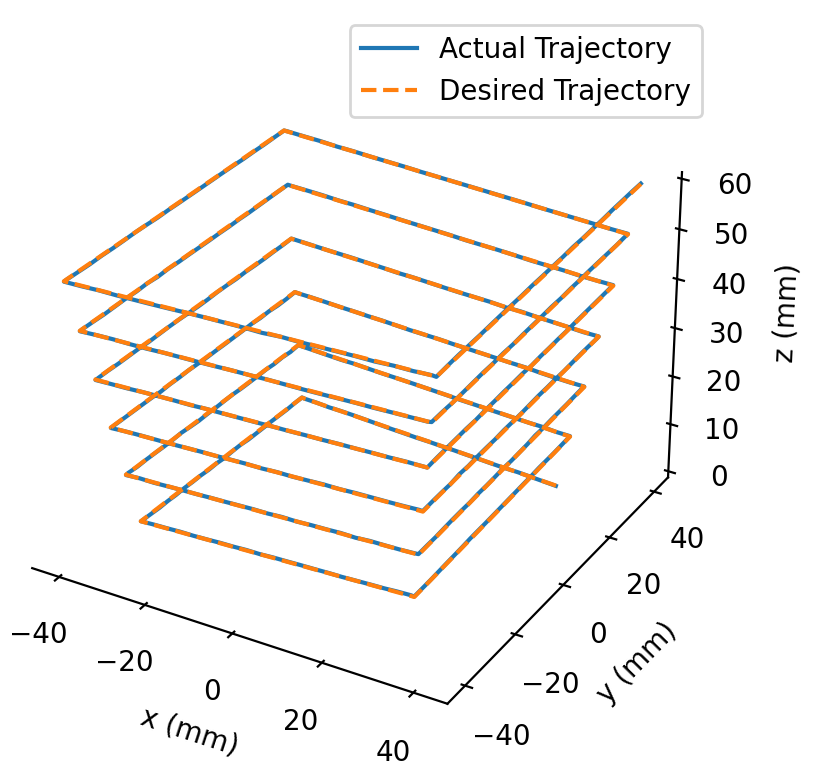}
\caption{Truncated pyramid tool tip trajectory with RCM constraint.}
\label{fig:pyramid}
\end{figure}

To evaluate the system's performance, extensive testing was conducted to determine how effectively the surgical tool remained constrained within the hole during teleoperation. Two distinct trajectories were employed for assessment, with the tool tip trajectory tracking error and RCM error serving as primary evaluation metrics. Trajectory tracking error measures the discrepancy between the actual tool tip position obtained from the robot and the desired path. Meanwhile, RCM error quantifies the difference between the initially recorded fulcrum point and the corresponding point calculated through inverse kinematics. Initially, truncated cone and truncated pyramid trajectories were examined on the digital twin platform to verify their feasibility and validate the RCM approach. Upon implementation on the real system, precise trajectory tracking was achieved, as demonstrated in Fig.~\ref{fig:cone} and Fig.~\ref{fig:pyramid}, while adhering to the RCM constraint. Maximum error and root mean square error (RMSE) for both trajectory tracking and RCM are provided in Table~\ref{table:error}.

\renewcommand{\arraystretch}{1.2}
\begin{table}[t!]
\centering
\caption{Evaluation of Tool Tip Trajectory Tracking with RCM Constraint.}
\begin{tabular}{|c|c|c|c|c|}
\hline 
\multirow{2}{*}{Trajectory (mm)} & \multicolumn{2}{c|}{Trajectory Tracking Error} & \multicolumn{2}{c|}{RCM Error}\\ \cline{2-5}
                                 &       \textbf{Max} & \textbf{RMSE}             &   \textbf{Max} & \textbf{RMSE}\\ \hline
Truncated Cone & 0.217108 & 0.087139 & 0.004 & 0.002 \\ \hline
Truncated Pyramid & 0.207951 & 0.091318 & 0.005 & 0.002 \\ \hline
\end{tabular}
\label{table:error}
\end{table}

For both trajectories, the maximum trajectory tracking error is approximately 0.2 mm, with the RMSE being less than 0.1 mm, demonstrating high accuracy in trajectory tracking. Regarding the RCM, the maximum error is kept below 5 $\mu$m, and the RMSE is merely 2 $\mu$m, highlighting remarkable precision in maintaining the RCM constraint. Teleoperation was also tested, revealing a maximum RCM error of 0.02 mm and an RMSE of 0.01 mm. These results indicate effective performance even under human control, offering a precise and safe training experience. 

\begin{figure}[thpb!]
\centering
\includegraphics[scale=0.3]{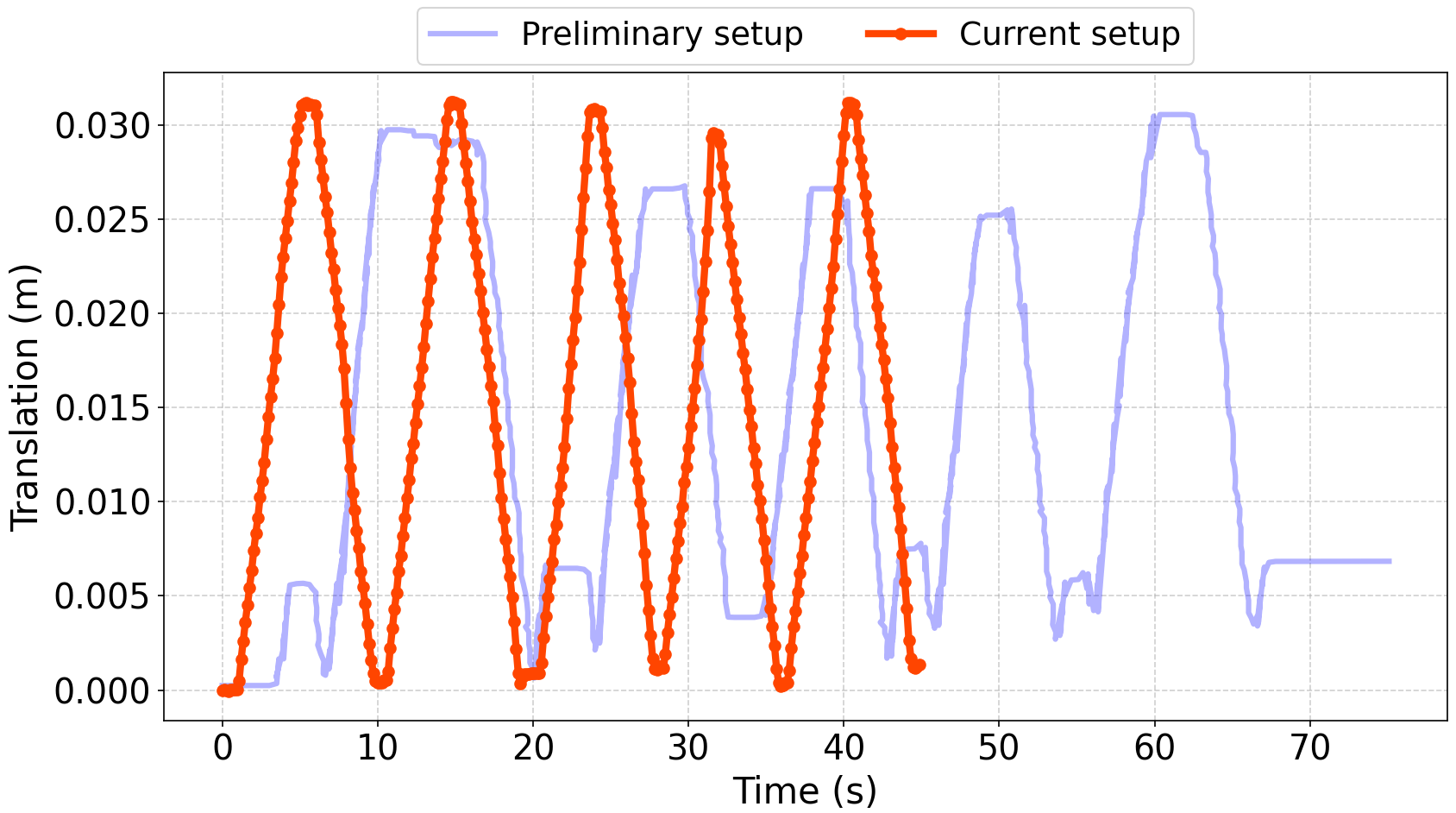}
\caption{Translation of the laparoscopic tool tip over time.}
\label{fig:Tool_tip_position}
\end{figure}

An experiment was conducted where the tool tip was pushed into foam material five times to measure its displacement over time. The results indicate that the current setup provides more stable, smoother, and faster teleoperation performance compared to the preliminary setup, as illustrated in Fig.~\ref{fig:Tool_tip_position}.

In the previous setup, the teleoperation of the UR3 occasionally exhibited non-smooth motion, including jerks, rumbling noises from the motors, slow robot movement, and delayed responses. Despite the limitations of the UR3 CB3 robot, which operates at a low-frequency rate of 125 Hz, these issues have been effectively addressed in the current development. By implementing a servo function using Real-Time Data Exchange (RTDE)\footnote{\href{https://sdurobotics.gitlab.io/ur\_rtde/}{https://sdurobotics.gitlab.io/ur\_rtde/}}, the system now incorporates a `lookahead time' parameter. This feature allows the robot to anticipate and adjust for future joint positions, ensuring smoother, faster, and more continuous transitions, effectively eliminating jerky movements. Notably, the system's latency has been reduced to just 0.01 seconds.

With the improved reaction time, a speed control slider was added to prevent excessive robot speeds during operation. In addition, robust safety features have been integrated, including continuous monitoring of the RCM error to ensure the tool remains constrained within the hole, as well as the avoidance of collisions and singularities. The system also monitors joint speeds and limits, prevents the robot from changing configuration, and tracks linear reachability. These enhancements ensure both the precision and safety of the robotic surgery training system. A demo showing a training task performed by the setup is available below\footnote{\href{https://youtu.be/2sbEkd3Yk04}{Supplementary Video: https://youtu.be/2sbEkd3Yk04}}.

\section{CONCLUSION}

This work aimed to develop a cost-effective robotic laparoscopy training system that replicates the functionality of high-end robotic surgery setups. Given the high cost and limited accessibility of robotic surgery systems, the goal was to build a system that offers widespread access for both on-site and off-site training, as well as research endeavors. To address the limitations of the preliminary setup, an innovative, low-cost robotic end-effector was designed to closely emulate fully-fledged systems and overcome previous design flaws. In addition, a digital twin platform was established, facilitating comprehensive simulation, testing, and real-time monitoring, thereby enhancing both the development and deployment phases. Furthermore, teleoperation with Touch haptic devices has been optimized, resulting in notable improvements in trajectory tracking while maintaining the RCM constraint. The system achieved an RMSE of 5 $\mu$m and reduced latency to just 0.01 seconds. These results showcase the system’s effectiveness in providing smooth, continuous, and rapid motion, along with a comprehensive set of safety features. Future work will focus on exploring the potential of incorporating visual and haptic feedback to further enhance trainee performance. Additionally, usability testing with surgeons will be conducted to gather feedback on the system's functionality and performance, aiming for a more intuitive and efficient user experience.

\addtolength{\textheight}{-12cm}  


\bibliographystyle{IEEEtran}
\bibliography{ref}
\end{document}